\def\year{2022}\relax

\documentclass[letterpaper]{article} 
\pdfoutput=1
\usepackage[T1]{fontenc}
\usepackage{aaai21}  
\usepackage{times}  
\usepackage{helvet}  
\usepackage{courier}  
\usepackage[hyphens]{url}  
\usepackage{graphicx} 
\usepackage{natbib}  
\usepackage{caption} 
\DeclareCaptionStyle{ruled}{labelfont=normalfont,labelsep=colon,strut=off} 
\usepackage{amsmath,amssymb}
\usepackage{algorithm}
\usepackage{algorithmicx}
\usepackage{makecell}

\newcommand{\GREEN}[1]{\textcolor[rgb]{0,1,0}{#1}}
\newcommand{\BLUE}[1]{\textcolor[rgb]{0,0,1}{#1}}
\newcommand{\YELLOW}[1]{\textcolor[rgb]{0.9,0.8,0}{#1}}
\usepackage{newfloat}
\usepackage{listings}
\floatstyle{ruled}
\newfloat{listing}{tb}{lst}{}
\floatname{listing}{Listing}
\usepackage{graphicx}

\usepackage{tikz}
\usepackage{comment}
\usepackage{amsmath,amssymb} 
\usepackage{color}
\usepackage{times}
\usepackage{epsfig}
\usepackage{graphicx}
\usepackage{bm}
\usepackage{multirow}
\usepackage{makecell}
\usepackage{bbding}
\usepackage{algpseudocode}
\usepackage{multirow}
\usepackage{booktabs}
\usepackage{tablefootnote}
\usepackage{threeparttable}
\usepackage{pifont}
 \usepackage[implicit=false]{hyperref}
\newcommand{\cmark}{\ding{51}}

\title{CNTN: Cyclic Noise-tolerant Network for Gait Recognition}
\author{Weichen Yu\textsuperscript{\rm 1}\footnote{This work is partly done in Watrix},
    Hongyuan Yu\textsuperscript{\rm 1},
    Yan Huang\textsuperscript{\rm 1},
    Chunshui Cao\textsuperscript{\rm 2},
    Liang Wang\textsuperscript{\rm 1}}
\affiliations{\\
	\textsuperscript{\rm 1} Center for Research on Intelligent Perception and Computing, Institute of Automation, Chinese Academy of Sciences\\
	\textsuperscript{\rm 2} Watrix.ai\\
	yuweichen16@mails.ucas.ac.cn,
	hongyuan.yu@cripac.ia.ac.cn,
	yhuang@nlpr.ia.ac.cn,
	chunshui.cao@watrix.ai,
	wangliang@nlpr.ia.ac.cn
	}

\begin{document}
\maketitle

\begin{abstract}
Gait recognition aims to identify individuals by recognizing their walking patterns. However, an observation is made that most of the previous gait recognition methods degenerate significantly due to two memorization effects, namely appearance memorization and label noise memorization. To address the problem, for the first time noisy gait recognition is studied, and a cyclic noise-tolerant network (CNTN) is proposed with a cyclic training algorithm, which equips the two parallel networks with explicitly different abilities, namely one forgetting network and one memorizing network. The overall model will not memorize the pattern unless the two different networks both memorize it.
Further, a more refined co-teaching constraint is imposed to help the model learn intrinsic patterns which are less influenced by memorization. Also, to address label noise memorization, an adaptive noise detection module is proposed to rule out the samples with high possibility to be noisy from updating the model.
Experiments are conducted on the three most popular benchmarks and CNTN achieves state-of-the-art performances. We also reconstruct two noisy gait recognition datasets, and CNTN gains significant improvements (especially 6\% improvements on CL setting). CNTN is also compatible with any off-the-shelf backbones and improves them consistently.

\end{abstract}

\section{Introduction}
Gait Recognition is a biometric technique that can be performed at a long distance without the subject’s cooperation, making it convenient for public safety and intelligent transportation \cite{KUMAR2021103052,bouchrika2011using}. And recent gait recognition methods \cite{9714177,huang20213d,chao2021gaitset,wu2016comprehensive,chen2021multi,zhang2021cross,li2022strong} have achieved encouraging progress.

However, previous gait recognition methods generally overlook two memorization effects that correspond to two characteristics of practical gait sequences respectively, which would lead to significant performance degeneration. 
The first is named `appearance memorization' since silhouette-based methods tend to memorize the outlines of the pedestrians, which are easily influenced by the appearances such as clothing conditions. Thus when the same pedestrian walks in different clothing, most existing gait methods memorize previous clothing patterns and degenerate severely. For example, in one of the most popular datasets CASIA-B \cite{6115889}, performances of previous state-of-the-art methods degrade by about 15\% (Tab.\ref{tab:sota_casiab}). The second is noisy label memorization. Given the fact that gait sequences are hard to annotate compared to other recognition tasks, human annotators make mistakes easily and introduce label noise. Also, recently unsupervised clustering methods are widely used to label the gait videos, which again introduces a problem of noisy datasets. Previous methods based on deep neural networks can easily memorize these label noise and show a significant performance degeneration \cite{algan2021image}. In Tab.\ref{tab:noisy}, the degeneration is more than 30\%. To the best of our knowledge, the problem of noisy gait recognition has not been investigated yet, however, is highly desired in realistic applications.

\begin{figure}[!t]
	\centering
	\includegraphics[width=0.45\textwidth]{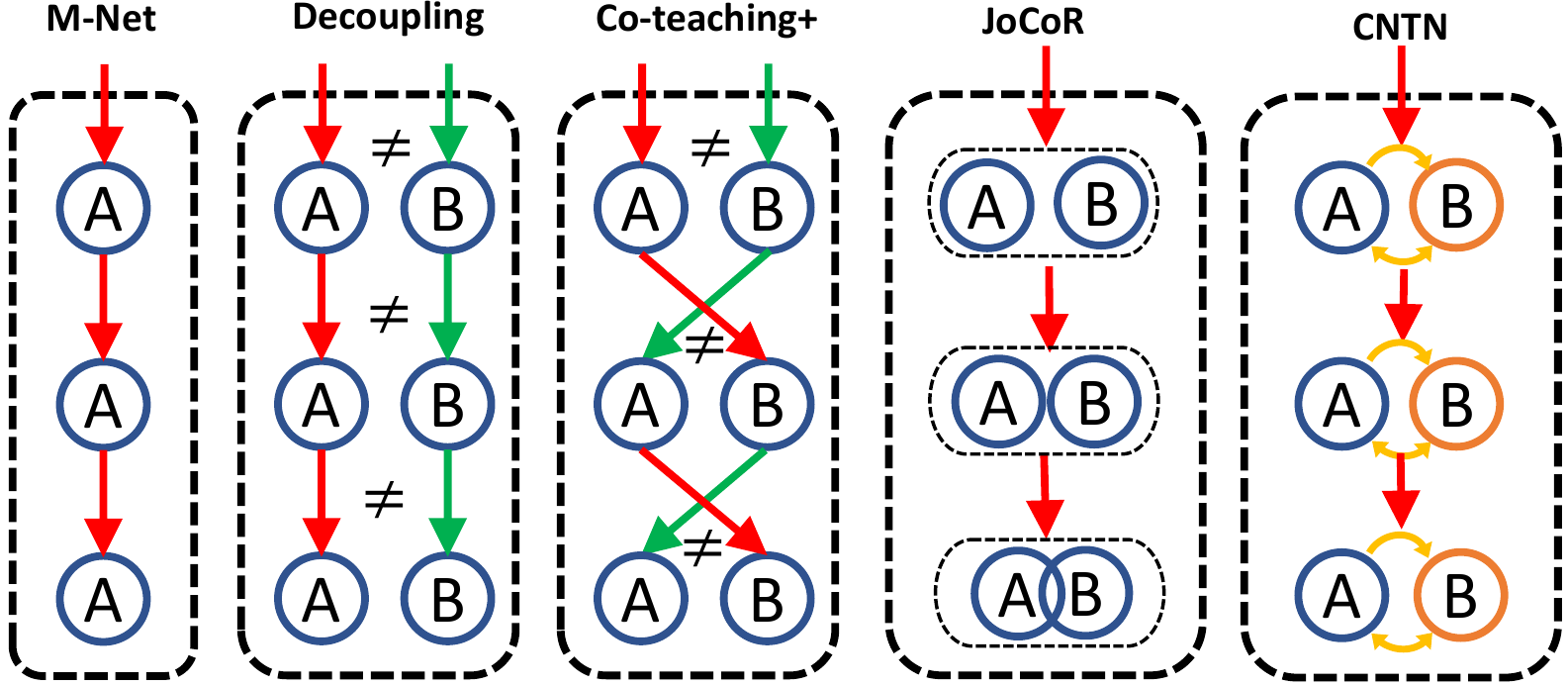}
	\caption{Comparison of algorithms among MentorNet (M-Net),
Decoupling, Co-teaching+, JoCoR and CNTN. (1) M-Net \cite{jiang2018mentornet} maintains only one network using meta learning. (2) Decoupling \cite{malach2017decoupling} updates the parameters of two networks when the predictions of them disagree.
(3) Co-teaching+ \cite{yu2019does} feeds forward and predicts each mini-batch data first, and keep prediction disagreement data only. (4) JoCoR \cite{wei2020combating} trains the two networks as a
whole and reduce the diversity of the two networks during training. (5) CNTN uses cyclic algorithm for training and explicitly design the two network to have different abilities.}
	\label{fig:co-teaching}
	\vspace{-2em}
\end{figure}


In this paper, we recast the gait recognition problem from the perspective of resisting memorization effects and propose a cyclic noise-tolerant network (CNTN) that decreases the model's memorization ability in threefold. Firstly, CNTN equips two parallel networks with explicitly different abilities, where the overall model will not memorize the pattern unless the two different networks both memorize it. Secondly, instead of learning purely supervised by labels, CNTN encourages the two networks to learn cyclically from each other, where the patterns learned are more related to the intrinsic characteristic of gait sequences themselves instead of memorizing the label-related patterns. Thirdly, to better address label noise memorization, CNTN also contains a module that prevents samples with high possibility to be noisy labels from updating the networks' parameters.

\begin{figure*}[!t]
	\centering
	\includegraphics[width=1.0\textwidth]{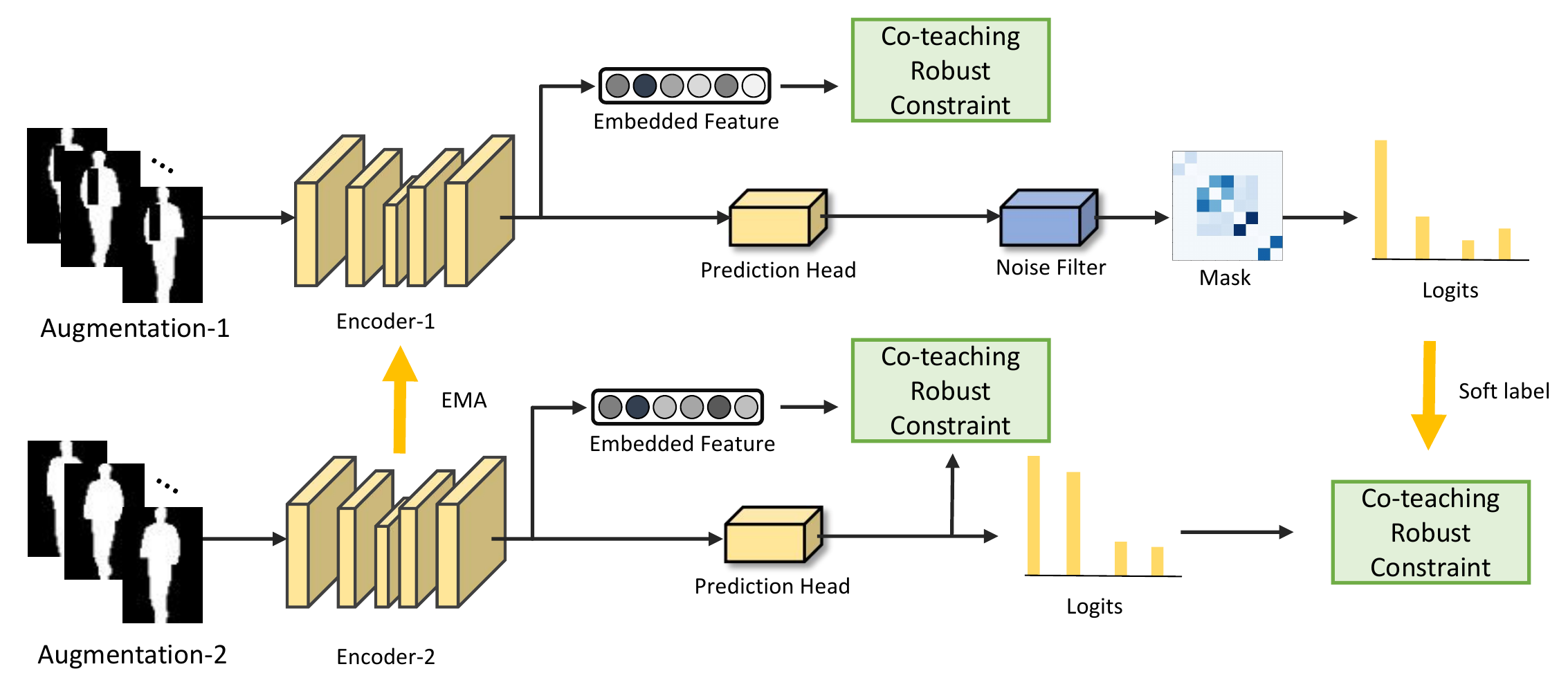}
 \vspace{-1em}
	\caption{Illustration of the proposed CNTN. CNTN contains two networks, the memorizing network (up) and the forgetting network (down). \GREEN{Green} modules represent the co-teaching robust constraint. \BLUE{Blue} module including noise filter and the mask represents the AND module. The memorizing network and forgetting network teach each other in a cyclical way (as in the \YELLOW{yellow} parts)}
	\label{fig:model}
	\vspace{-2em}
\end{figure*}

More specifically, CNTN includes a cyclic noise-tolerant algorithm (CNTN algorithm), a co-teaching robust constraint (CRC), and an adaptive noise detection (AND) module. CNTN algorithm performs per iteration and is able to keep two networks different: not collapse into one single network even after numerous finite iterations of training. The differences between the networks are essential for resisting both appearance and label noise memorization effects. Moreover, the two parallel networks, namely a forgetting network and a memorizing network co-teach each other in a cyclic way (Fig.\ref{fig:co-teaching} and Fig.\ref{fig:model}, yellow parts), and CRC is computed by the logits from the two networks for mining intrinsic patterns. CRC is designed to involve `self-supervised' learning between different outlines of the same identity, which decreases the model's appearance memorization and one term in CRC is not directly dependent of labels, which decreases the model's noise label memorization. AND module detects possibly noisy samples based on their entropy, and prevent these samples from leading the network to memorize noisy labels.

To verify the memorization effects of CNTN, experiments are conducted on both the popular gait recognition benchmarks and the two reconstructed gait recognition noisy datasets. On the three public benchmarks, CNTN
achieves state-of-the-art performances. And the noisy reconstructed datasets include both appearance memorization settings and label noise memorization settings. Note that CNTN is model-agnostic and improve the performances regardless of different backbones.

Our main contributions are summarized as below:
\begin{itemize}
	\item To the best of our knowledge, for the first time, the problem of noisy gait recognition is studied, which is in great demand in real applications.
	\item  A novel memorization-alleviating gait recognition framework named CNTN is proposed, which contains a cyclic noise-tolerant algorithm, a forgetting network and a memorizing network. And the co-teaching robust constraint is imposed to refine co-teaching and decrease memorization. An adaptive noise detection module based on entropy is proposed to better address label noise memorization.
	\item Extensive experiments are conducted to illustrate the effectiveness of CNTN. On noisy datasets including two practical settings, there is a significant gain (especially 6\% improvements on CL setting). And on the three public benchmarks, CNTN achieves state-of-the-art performances. Note that CNTN is model-agnostic and improve the performances consistently.
\end{itemize}

\section{Related Works}
\subsection{Gait Recognition}
Gait recognition \cite{sarkar2005humanid,wu2016comprehensive,pan2020optimization,Li_2020_ACCV,goffredo2009self,liu2011joint,jean2009towards} is to learn the unique spatio-temporal pattern about the human gait characteristics and obtain the identity information. The gait model can be categorized into two types \cite{kusakunniran2013recognizing,hu2013view}. The first is 3D skeleton models which are reconstructed from different camera views \cite{ariyanto2011model,bodor2009view,zhao20063d}, while the other is 2D gait data including silhouettes \cite{lin2021gait,hou2020gait,chao2021gaitset} and gait energy image (GEI) \cite{shiraga2016geinet,zhang2016siamese}. This paper focuses on the widely used 2D silhouette gait input.
Refined spatial-temporal networks are designed recently. Zhang et al. \cite{zhang2019cross} utilizes a temporal attention mechanism on each frame and adaptively adjusts the weights of different frames. Wolf et al. \cite{wolf2016multi} first utilize 3D convolution for feature extraction in gait recognition. Zhang et al. \cite{zhang2020learning} emphasizes disentangled representation learning. Further works \cite{chen2021multi,yu2017gaitgan} also utilize generated adversarial networks to generate more data to help with training. Self-supervised \cite{9413894} is also used to learn to perform gait recognition. More works \cite{zhu2021gait,zhang2022realgait} focus on real gait conditions and provide new datasets.
However, existing works lack a high capacity to prevent memorization effects and exhibit a severe degeneration when encountering label noise and clothing change. 
\subsection{Noisy label learning}
Label noise is common in real-world datasets, due to crowd-sourcing, data acquired on the web and social media. 
Noisy labels learning (NLL) is widely studied and has sophisticated categories in some surveys \cite{songlearning,algan2021image}. We classify these methods into three types. The first category \cite{jiang2018mentornet,lee2018cleannet,jaehwan2019photometric} utilizes a re-annotated subset to help obtain noise-identifying ability. The second category doesn't require a clean subset, but requires assumptions or prior knowledge on noise pattern and noise rate, and tends to add additional noise adaption layers to implicitly or explicitly construct noise transition matrix \cite{goldbergertraining,patrini2017making}. The third type needs neither clean sets nor noise knowledge, which mainly focuses on regularization methods \cite{zhang2018mixup}, robust loss design \cite{wang2019symmetric}, meta-learning \cite{algan2020meta}, etc.
However, prior knowledge of noise rate and noise type or clean subsets are not always practical in real applications.


\section{Methodology}

\subsection{Preliminary} \label{Sect:Preliminary}
Previous gait recognition methods perform well on normal gait datasets, but degenerate intensively on noisy gait datasets, which is mainly due to the deep neural networks' (DNN) memorization effects. Memorization effects is a fundamental problem but is neglected before in gait recognition, and we will first introduce several empirical phenomena/principles concluded from excellent previous works \cite{goodfellow2016deep,zhang2021understanding,pmlr-v70-arpit17a}. 1) \textbf{Memorization effect}. DNN has a high capacity to overfit identity-irrelevant knowledge and can fit even unstructured random noise without substantially longer training time \cite{zhang2021understanding,arpit2017closer}. This is named as memorization effect since the DNN tends to `memorize' all samples instead of learning useful patterns. 2) Another empirical phenomenon is \textbf{degeneration effect}, which states that DNNs tend to learn clean and easy patterns first and then gradually fit noisy labels later, and the decision boundaries become more complicated \cite{pmlr-v70-arpit17a,zhang2021understanding}. 3) Due to the degeneration effect, methods in noisy label learning widely uses \textbf{small loss trick}, which is to treat samples with small training loss as clean ones \cite{chen2019understanding,jiang2018mentornet,song2019does}.

\subsection{Cyclic Noise-tolerant Network}
CNTN is inspired by co-teaching and co-training in image classification tasks \cite{han2018co,yu2019does,chen2021boosting,xiang2021learning,malach2017decoupling,chen2019understanding,wei2020combating} which utilize two parallel networks for resisting noise. Co-teaching has shown that two parallel networks have different abilities to learn and can filter out different noisy data \cite{han2018co,chen2019understanding}. In this paper, CNTN is proposed to utilize the principle of co-teaching to resist both appearance and label noise memorization effects because the model will not memorize a certain appearance pattern or noise pattern unless the two networks both memorize it. However, previous co-teaching methods have three problems which make them inappropriate for gait recognition task.

1) Co-teaching in other tasks \cite{han2018co,yu2019does,chen2019understanding} needs prior knowledge about the noise rate or noise type for selecting clean samples, which is not always available in gait recognition settings since even human annotators often make mistakes. 2) Previous co-teaching in image recognition tasks \cite{han2018co,yu2019does,chen2021boosting,malach2017decoupling,chen2019understanding,wei2020combating} impose the differences between the two networks by different initialization. However, in gait recognition task gait features have small differences between classes \cite{yu2022generalized,lin2020gait}, after numerous epochs of training, the two networks gradually converge to each other and their differences are weakened, which collapse into a single model and the memorization ability increases. 
3) Most previous works exchange information in sample-level, which wastes useful gait information in more refined features.

To address the problems above, we propose CNTN which does not require any prior knowledge about noise rate or noise pattern. By carefully designing the algorithm, the different abilities between two networks can be guaranteed: one for memorizing stable patterns and the other for learning new patterns. The two networks are asymmetric, more discriminative and have respective memorization patterns. Besides, we impose more refined knowledge exchange in gait features, which keeps more information and are more computationally efficient.

{\setlength\abovedisplayskip{0cm}
\setlength\belowdisplayskip{0cm}
\begin{algorithm}[tb]
    \vspace{-0.1cm}
\caption{CNTN Algorithm}
\label{alg}
\textbf{Input}: Batch size $N$, structure of network $f$, a data augment function $t \sim \mathcal{T} $, momentum $m$.
\begin{algorithmic}[1]
\State Initialize $f$ with randoms and get $f_f$.
\State Initialize $f$ with randoms and get $f_m$.
\For{sampled minibatch $\{(x_i,y_i)\}_{i=1}^N$}
\State $\mathcal{L} = 0$
\State $f_m = m f_m$ + $ (1-m)f_f$
\For {$i \in \{1,,N\} $}
\State $z_i^f, p_i^f = f_f(t(x_i))$
\State $z_i^m, p_i^m = f_m(t(x_i))$
\State Calculate $\mathcal{L}$ +$=\mathcal{L}_{c}$, according to Eq.~\ref{eq:coteaching-loss}
\EndFor
\State $\mathcal{L} $ +$= \mathcal{L}_{s}$
\State update $f_f$,$f_m$
\EndFor
\end{algorithmic}
\textbf{Output}:$f_f$
\end{algorithm}}

\subsubsection{Cyclic Noise-tolerant Algorithm}
As shown in Fig.\ref{fig:model} and Alg.\ref{alg}, the memorizing network ($M$, the upper one) and forgetting network ($F$, the one below) teach each other in every iteration cyclically: 1) at the start of an iteration, $F$ passes its network parameters to $M$ by using exponential moving average (EMA); 2) both the networks are given with the same input sequences $x_i$ with a sampled data augmentation method $t \sim \mathcal{T} $; 3) $M$ and $F$ takes $t_m(x_i)$ and $t_f(x_i)$ as input and after the encoder, they output feature representations $x_i^f$, $x_i^m$; 4) after the prediction head, the predicting probability of $M$: $p_i^m$is used as soft pseudo-label to supervise $F$: $p_i^f$, as in Eq.\ref{eq:coteaching-loss}; 5) $F$ is updated by triplet and cross-entropy (CE) loss while $M$ isn't using CE or triplet loss. An iteration ends.

In every iteration, at first $F$ serves as a teacher and passes knowledge to $M$, while $M$ learns from $F$ in an accumulative manner. Later, $M$ serves as a teacher and provide a generated soft-label (as in Eq.\ref{eq:coteaching-loss}) for $F$, while $F$ learns from $M$ about the past stable patterns. At the back-propagation, the loss term contains pattern knowledge from both networks and $F$ and $M$ are updated using different gradients.
\subsubsection{Co-teaching Robust Constraint (CRC)} In previous works \cite{han2018co,yu2019does,chen2019understanding}, the information exchange lays mainly in sample exchange, where one network passes the samples it considers clean to another. However, we argue that there are three problems.

1) Refined knowledge exchange. The sample-level information exchange waste the knowledge such as how confident network A believes the sample is clean, whether there are other classes that network A considers the sample may belong to with relatively high possibility. Thus, we propose to exchange in logits level which is more refined, as in Eq.\ref{eq:coteaching-loss}.
2) Intrinsic patterns. Previous methods mostly includes the given labels in the constraints, while CRC also includes a term of consistency between the two networks, which encourages the model to learn intrinsic patterns and less susceptible to label noise.
3) High efficiency. Previous sample exchange requires the other network to forward propagate the clean samples once more for acquiring loss value, while in the proposed constraint the loss is computed only once. In co-teaching, assume every mini-batch contains $N$ samples and the noise rate is $\sigma$ ($\sigma$ is not needed for the proposed algorithm), then the estimated computing cost is $2N(2-\sigma)$ for co-teaching and $2N$ for the proposed method with data augmentation and $N$ without data augmentation. Since usually the $\sigma$ is small, the proposed method accelerates the forward propagation by about 80\% to 400\%.

\begin{equation} \label{eq:coteaching-loss}
{L_c} = -Softmax(p_m) log Softmax(p_f)
\end{equation}
where $p_f$ is the prediction from forgetting network and $p_m$ is the prediction from memorizing network.
\begin{table*}[th]
  \scriptsize
  \centering
   \caption{Rank-1 accuracy (\%) on Noisy Clothing CASIA-B under all views, different conditions, excluding identical-view case.
  }
   \vspace*{-1em}
  \resizebox{0.98\textwidth}{!}{
    \begin{tabular}{c|c|c|c|c|c|c|c|c|c|c|c|c|c}
    \toprule
    \multicolumn{2}{c|}{Gallery NM\#1-4}  &\multicolumn{12}{c}{$0^{\circ}$-$180^{\circ}$} \\
    \hline
     \multicolumn{2}{c|}{Probe} &   $0^{\circ}$     & $18^{\circ}$    & $36^{\circ}$    & $54^{\circ}$    & $72^{\circ}$    & $90^{\circ}$    & $108^{\circ}$   & $126^{\circ}$   & $144^{\circ}$   & $162^{\circ}$   & $180^{\circ}$  & Mean\\
    \midrule

    \multicolumn{1}{c|}{\multirow{1.5}[2]{*}{NM\#5-6}}  & Baseline &  91.1    &  98.3     &  99.0    & 98.0  &   94.5    & 91.8  &   94.4    & 98.0  &   98.8   &   96.2    &  90.4    & 95.5 \\
     & \textbf{Ours} & \textbf{94.0}& \textbf{98.8}& \textbf{99.1} &\textbf{98.5}& \textbf{95.9}& \textbf{94.6}& \textbf{96.5}& \textbf{98.4}& \textbf{98.6} &\textbf{98.4} &\textbf{93.4} & \textbf{96.9} \\

     \cline{1-14}  \multicolumn{1}{c|}{\multirow{1.5}[2]{*}{BG\#1-2}} & Baseline & 88.2& 93.9& 95.5& 93.7 &90.0& 85.0 &89.9& 94.0 &94.5& 93.9& 85.5&91.3  \\

      & \textbf{Ours}  & \textbf{91.8} & \textbf{94.2} & \textbf{95.6} & \textbf{94.6} & \textbf{91.0} & \textbf{87.4} & \textbf{89.3} & \textbf{95.2} & \textbf{96.2} & \textbf{94.1} & \textbf{87.7} & \textbf{92.5} \\

\cline{1-14} \multicolumn{1}{c|}{\multirow{1.5}[2]{*}{CL\#1-2}}   & Baseline  & 46.3&  56.2&  62.8&  57.7&  54.3&  56.3&  54.1&  58.2&  58.9&  53.4&  40.5&  54.4  \\

      & \textbf{Ours}  & \textbf{47.4} & \textbf{64.3} & \textbf{67.2} & \textbf{65.6} & \textbf{61.1} & \textbf{60.7} & \textbf{62.6} & \textbf{65.6} & \textbf{65.1} & \textbf{58.9} & \textbf{44.6} & \textbf{60.3} \\

    \bottomrule

    \end{tabular}%
    }

\label{tab:noisy}
\vspace*{-2em}
\end{table*}%
Apart from the proposed constraints in Eq.\ref{eq:coteaching-loss}, the CRC also includes a modified supervised loss in Eq.\ref{eq:nceloss} for more robust feature learning. The triplet loss and cross entropy loss are traditional and just follows previous works \cite{chao2021gaitset,fan2020gaitpart,KUMAR2021103052}.
{\setlength\abovedisplayskip{0cm}
\setlength\belowdisplayskip{0cm}
\begin{equation} \label{eq:nceloss}
\begin{split}
{L_{MIL}} &= -log {\frac{\sum_{k^+}e^{qk^+}}{\sum_{k^+}e^{qk^+}+\sum_{k^-}{e^{qk^-}}}}
\end{split}
\end{equation}}
where $q$ is the query, $k^+$ is the representation of the same class and $k^-$ is the representation of different class of $q$.
{\setlength\abovedisplayskip{0cm}
\setlength\belowdisplayskip{0cm}
\begin{equation} \label{eq:total loss}
{L_{CRC}} = \sigma_0L_c+\sigma_{1}L_{CE}+\sigma_{2}L_{tri}+\sigma_{3}L_{MIL}
\end{equation}}
We add a term of multi-sample InfoNceLoss (MIL) (Eq.\ref{eq:nceloss}) to further address the memorization effect for the following reason. Traditional triplet loss uses one anchor-positive pair and one anchor-negative pair at a time (one-to-one). And InfoNceLoss, usually used in self-supervised training \cite{chen2020improved}, uses one anchor-positive pair and many anchor-negative pairs at a time (one-to-many). Inspired by supervised contrastive loss \cite{NEURIPS2020_d89a66c7}, more pairs in the numerator and the dominator help with better feature distribution in high dimensional space. We argue that many-to-many loss term also improves the resistance of memorization effect. If one noisy data exists in triplet loss term, the one-to-one loss value is greatly influenced. On the other hand, if one noisy sample exists in MIL term, the many-to-many loss value is influenced but to a smaller degree, which is more resistant to noisy data.
\subsubsection{Mathematical Analysis - CNTN}
In this part, we aim to view CNTN mathematically about how the memorizing network differs from the forgetting network even after a large finite number of epochs.
{\setlength\abovedisplayskip{0cm}
\setlength\belowdisplayskip{0cm}
\begin{equation}
\theta_N^f = \theta_0^f+\sum_{k=1}^{N}{\Delta\theta_k^f}
\end{equation}}
{\setlength\abovedisplayskip{0cm}
\setlength\belowdisplayskip{0cm}
\begin{equation} \label{eq:update}
\begin{split}
\theta_N^m &= \theta_0^f+\sum_{k=1}^{N-1}{\Delta\theta_k^f}+m^N(\theta_0^m-\theta_0^f)\\&+\sum_{k=1}^{N-1}{m^{N-k}}{(\Delta\theta_k^m-\Delta\theta_k^f)} + \Delta\theta_N^m \\
&= \theta_0^f + m^{N}(\theta_0^m-\theta_0^f)\\&+\sum_{k=1}^N{m^{N-k}\Delta\theta_k^m+{(1-m^{N-k})\Delta\theta_k^f}}
\end{split}
\end{equation}}
By utilizing CNTN, the network parameters can be represented as $\theta_0^f$ and $\theta_0^m$ at initialization $t_0$. And in iteration $k$ we update the forgetting and memorizing networks with $\Delta\theta_k^f$ and $\Delta\theta_k^m$. Then we can deduce from Alg.\ref{alg} that at iteration N, their parameter can be as follows:

When the training iteration $N$ is set, and momentum $m<1$, the weight of the co-teaching $m^{N-k}$ increases as $k$ increases, and the weight of the supervised learning $1-m^{N-k}$ decreases as $k$ increases. When $N$ is large, $m$ is close to 1 \footnote{m is 0.99 in this paper.}, the second item $m^{N}(\theta_0^m-\theta_0^f)$ decays to near zero. What's really important here is the third term in Eq.\ref{eq:update}. Note that for the memorizing network, as in Fig.\ref{fig:model}, it is not using the traditional supervised triplet loss or cross-entropy loss to update its weights, thus the term $\Delta\theta_k^m, k \in \{1,,,N\}$ is directly dependent on the consistency constraints in Eq.\ref{eq:coteaching-loss}, which aims to learn the intrinsic data patterns. And $\Delta\theta_k^f, k \in \{1,,,N\}$ is mainly dependent on supervised learning.

The well-designed CNTN algorithm has its advantages: 1) As in the above equations, the memorizing network is able to adjust the percentage of traditional supervised (CE and triplet loss) and co-teaching automatically; 2) The term $(1-m^{N-k})\Delta\theta_k^f$ which comes from the process that $F$ transfer knowledge to $M$ helps the whole two networks converge; 3) As in the Section \ref{Sect:Preliminary} the degeneration states, CNTN algorithm utilizes this effect and the $M$ network keeps more previous knowledge, which is considered to be more robust and more clean patterns. 4) The term $m^{N-k}\Delta\theta_k^m$ guarantees that $M$ is different from $F$ even after large finite epochs. And the differences between them are the key point for co-teaching, making these two networks have different abilities: one is good at utilizing supervised labels and adapts to new but noisy knowledge fast, and another is good at mining intrinsic data similarities and keeping stable representations.
\\\textbf{\textit{Differences Between CNTN And Fast-slow Methods}}. Fast-Slow methods \cite{feichtenhofer2019slowfast,zhao2021mgsvf} also have two parallel networks, however they are very different in principle. Slowfast network \cite{feichtenhofer2019slowfast} uses information at two different frame rate in video recognition, and Mgsvf \cite{zhao2021mgsvf} is frequency-awared inter and intra-space in few-shot class-incremental learning, learns projection to another feature $f:Z_1\gets Z_0$ space and constrains the feature representation to be close in two neighboring timestamp, $L_{speed} = ||z_j^{(t)}-z_j^{(t-1)}||$.
\\\textbf{\textit{Differences Between CNTN And Other Co-teaching Methods}}. As shown in Fig.\ref{fig:co-teaching}, there are three main differences. 1) previous frameworks are not cyclic; 2) the two networks are not explicitly made different memorization abilities; 3) information exchange is of different granularity.

\begin{table}[t]
	\centering
	\caption{{Component-wise analysis. Ablation study on Noisy Clothing CASIA-B.}}
  \vspace*{-1em}
     \setlength{\tabcolsep}{0.9mm}{
		\begin{tabular}{c|c|cccc|ccc}
		
			\hline
			&  {\multirow{1}{*}{}}  & $L_s$ & $L_{\small{c}}$ & \makecell[c]{Cyclic\\ Training}   & AND &  NM & BG &CL \\
			 \hline
			 \#1 & \multicolumn{1}{c|}{\multirow{3}[2]{*}{Supervised}} & \cmark & & & & 95.5  & 91.3 & 54.4 \\
			
			 \#2 & &\cmark  &  & \cmark  & &  96.6  & 91.9 & 56.7 \\
			 \#3 & &\cmark  &  & \cmark  & \cmark &  96.8  & 92.1 & 57.9 \\

			 \hline
			 \#4& \multicolumn{1}{c|}{\multirow{1.5}[2]{*}{\makecell[c]{Self- \\ supervised}}} &  & \cmark &  & & 9.1  & 6.3 & 3.6 \\
			 \#5 & & &\cmark & \cmark   &  & 15.1  & 9.8 & 5.3 \\
			\hline
			\#6 & \multicolumn{1}{c|}{\multirow{3}[2]{*}{Ours}} &\cmark  & \cmark &  & &  96.8  & 91.9 & 57.7 \\
			\#7 & &\cmark  & \cmark & \cmark  & &  \textbf{97.2}  & 92.4 & 59.7 \\
			\#8 & &\cmark  & \cmark & \cmark  & \cmark &  96.9  &\textbf{92.5} &\textbf{60.3} \\
			\hline
		\end{tabular}
	}\label{tab:ablation}
	\normalsize
 \vspace*{-1em}
\end{table}
\begin{table}[t]
	\centering
	\caption{Model agnostic results on CASIA-B dataset.}
  \vspace*{-1em}
	\label{tab:agnostic}
		\setlength{\tabcolsep}{0.9mm}{
			\begin{tabular}{lcccc}
				\toprule[1.1pt]
				Methods  & NM $\bm{\uparrow}$ & CL  $\bm{\uparrow}$ & BG    $\bm{\uparrow}$  & Model Size
				\\ \midrule[1.1pt]
				GaitSet &$97.4$ & $93.7$ & $77.8$ & 2.59M \\
				\textbf{+Ours} & \textbf{97.5} & \textbf{94.3} & \textbf{79.1} & 2.59M  \\
				\midrule
				GaitGL & $97.4$ & $94.5$ & $83.6$ & 3.10M \\			
				\textbf{+Ours} & \textbf{97.5} & \textbf{94.8} & \textbf{86.1} & 3.10M \\
				\bottomrule[1.1pt]
		\end{tabular}} \label{tab:Model agnostic}
   \vspace*{-2em}
\end{table}

\begin{table*}[t]
  \centering
  \caption{Rank-1 accuracy (\%) on OUMVLP under 14 probe views, excluding identical-view cases.}
      \vspace*{-1em}
  \resizebox{0.98\textwidth}{!}{
    \begin{tabular}{c|c|c|c|c|c|c|c|c|c|c|c|c|c|c|c}
    \toprule
    \multirow{2}[2]{*}{\textbf{Method}} & \multicolumn{14}{c|}{\textbf{Probe View}}                                                  & \multicolumn{1}{c}{\multirow{2}[2]{*}{\textbf{Mean}}} \\
\cline{2-15}    \multicolumn{1}{c|}{} & $0^{\circ}$  & $15^{\circ}$  & $30^{\circ}$  & $45^{\circ}$  & $60^{\circ}$  & $75^{\circ}$  & $90^{\circ}$ & $180^{\circ}$  & $195^{\circ}$  & $210^{\circ}$  & $225^{\circ}$  & $240^{\circ}$  & $255^{\circ}$  & $270^{\circ}$  &  \\
    \midrule
    GEINet & 23.2  & 38.1  & 48.0  & 51.8  & 47.5  & 48.1  & 43.8  & 27.3  & 37.9  & 46.8  & 49.9  & 45.9  & 45.7  & 41.0  & 42.5  \\
    \hline
    GaitSet & 79.3  & 87.9  & 90.0  & 90.1  & 88.0  & 88.7  & 87.7  & 81.8  & 86.5  & 89.0  & 89.2  & 87.2  & 87.6  & 86.2  & 87.1  \\
    \hline
    GaitPart & 82.6  & 88.9  & 90.8  & 91.0  & 89.7  & 89.9  & 89.5  & 85.2  & 88.1  & 90.0  & 90.1  & 89.0  & 89.1  & 88.2  & 88.7  \\
    \hline
    GLN   & 83.8  & 90.0  & 91.0  & 91.2  & 90.3  & 90.0  & 89.4  & 85.3  & 89.1 & 90.5 & 90.6 & 89.6  & 89.3  & 88.5  & 89.2  \\
    \hline
    GaitGL  & 84.9 & 90.2 & 91.1 & 91.5 & 91.1 & 90.8 & 90.3 & 88.5 & 88.6  & 90.3  & 90.4  & 89.6 & 89.5 & 88.8 & 89.7 \\
    \hline
    Ours  & \textbf{87.2} & \textbf{90.9} & \textbf{91.4} & \textbf{91.6} & \textbf{91.4} & \textbf{91.1} & \textbf{90.9} & \textbf{90.2} & \textbf{89.7} & \textbf{90.5}  & \textbf{90.6}  & \textbf{90.2} & \textbf{90.0} & \textbf{89.7} & \textbf{90.4} \\
    \bottomrule
    \end{tabular}%
    }
    \vspace*{-2em}
  \label{comparision_oumvlp}%
\end{table*}%
\begin{table}[h]
\centering
\caption{ Rank-1 accuracy (\%) on Noisy-CASIA-B.} 
    \vspace*{-1em}
\begin{tabular}{|c|c|c|c|c|}
\hline
Noise Type                                         & Method        & NM            & BG            & CL            \\ \hline
\multicolumn{1}{|c|}{\multirow{2}{*}{\makecell[c]{Random noise \\ Noise rate=0.1}}} & Baseline      & 94.0          & 87.6         & 62.4          \\ \cline{2-5}
\multicolumn{1}{|c|}{}                             & \textbf{Ours} & \textbf{94.9} & \textbf{87.9} & \textbf{63.4} \\ \hline
\multicolumn{1}{|c|}{\multirow{2}{*}{\makecell[c]{Random noise \\ Noise rate=0.2}}} & Baseline      & 90.8          & 83.0          & 53.0          \\ \cline{2-5}
\multicolumn{1}{|c|}{}                             & \textbf{Ours} & \textbf{91.2} & \textbf{84.7} & \textbf{55.3} \\ \hline
\multirow{2}{*}{\makecell[c]{Augmentation noise \\ Noise rate=0.1}}                        & Baseline      & 94.5          & 88.7          & 67.7          \\ \cline{2-5}
& \textbf{Ours} & \textbf{95.1} & \textbf{89.5} & \textbf{68.2} \\ \hline
\multirow{2}{*}{\makecell[c]{Augmentation noise \\ Noise rate=0.2}}                        & Baseline      & 93.8          & 87.5          & 64.0          \\ \cline{2-5}
& \textbf{Ours} & \textbf{95.0} & \textbf{88.8} & \textbf{66.6} \\ \hline
\end{tabular}
    \vspace*{-2em}
\label{tab:random casiab}
\end{table}

\subsection{Adaptive Noise Detection Module}
To better address the noisy label memorization, we propose an Adaptive Noise Detection Module (AND) to filter out probable noisy data labels and provide relative clean gradients. Given that the entropy indicates the degree of whether the network is certain about the class the current data point belongs to, AND selects the labels with low entropy and high consistency.

Given the degeneration effect in Sec. 3.1, the data $x_i$ with the noisy label '074' and actually clean label `001' firstly is close to `001' class since they are similar in representation in the early training period. Later, due to memorization effect the network gradually memorize this noisy pattern and `001' is pushed away. During this period, it undergoes when it is confused about which class to belong to, and the prediction probability is even among two or more classes. Thus, we propose to select the ones with sharp class distribution and small cross-entropy loss. This principle accords with the "small loss trick" used by other noisy label tasks. In this way, the noisy patterns are resisted from updating the networks' parameters and alleviate the memorization effect.

\section{Experiments}

\subsection{Dataset}
We conduct experiments on public datasets CASIA-B \cite{6115889}, Outdoor-Gait \cite{song2019gaitnet} and OUMVLP \cite{takemura2018multi}.
We also reconstruct noisy gait datasets Noisy-CASIA-B and Noisy-Outdoor-Gait based on CASIA-B and Outdoor-Gait on both normal noisy settings as in noisy CIFAR \cite{han2018co} and noisy clothing setting since it is more close to practical gait applications. More details are in Appendix A.

\subsection{Experiments on Noisy Datasets}
\textbf{Backbone} GaitSet \cite{chao2019gaitset} has been taken as backbone by several works \cite{hou2020gait} and is also an important baseline in our experiments. GaitSet treats the silhouettes of a gait sequence as an unordered set and splits the features horizontally to learn part representations for gait recognition.
\\\textbf{CASIA-B} Table.\ref{tab:noisy} and Tab.\ref{tab:random casiab} shows the performance comparison between our methods and baseline on Noisy Clothing Casia-B and Noisy-CASIA-B. The probe sequences are divided into three subsets, i.e. NM, BG, CL, which are respectively evaluated. The accuracy for each probe view is averaged on all gallery views excluding the identical-view cases.

The following observations can be made: 1) The proposed framework shows excellent performance in datasets with hard clothing and noisy labels. In all cases, the margins over the baselines are significant, which indicates that the framework's ability in alleviating appearance and label noise memorization effect; 2) Although noisy clothing setting is harder and more practical and baseline performs poorly at CL, CNTN outperforms baseline CL 5.9\%. However, there is a still big margin between CNTN and clean datasets, showing that the problem of the noisy label is not fully solved.

\begin{table*}[!h]
  \scriptsize
  \centering

   \caption{Rank-1 accuracy (\%) on CASIA-B under all views, different settings and conditions, excluding identical-view case.}
    \vspace*{-1em}
    \begin{tabular}{c|c|c|c|c|c|c|c|c|c|c|c|c|c|c}
    \toprule
    \multicolumn{3}{c|}{Gallery NM\#1-4}  &\multicolumn{12}{c}{$0^{\circ}$-$180^{\circ}$} \\
    \hline
    \multicolumn{3}{c|}{Probe}    & $0^{\circ}$     & $18^{\circ}$    & $36^{\circ}$    & $54^{\circ}$    & $72^{\circ}$    & $90^{\circ}$    & $108^{\circ}$   & $126^{\circ}$   & $144^{\circ}$   & $162^{\circ}$   & $180^{\circ}$  & Mean\\
    \midrule

    \multicolumn{1}{c|}{\multirow{12}[2]{*}{\textbf{ST(24)}}} & \multicolumn{1}{c|}{\multirow{3}[2]{*}{NM\#5-6}}
       & GaitSet & 71.6  & 87.7  & 92.6 & 89.1  & 82.4  & 80.3 & 84.4  & 89.0  & 89.8  & 82.9  & 66.6  & 83.3  \\
&       & GaitGL  & 77.0 & 87.8 & 93.9 & 92.7 & 83.9 & 78.7 & 84.7 & 91.5 & 92.5 & 89.3 & 74.4 & 86.0 \\
&       & Ours  & \textbf{80.4} &  \textbf{91.2} &  \textbf{96.0} &  \textbf{93.0} &  \textbf{85.2} &  \textbf{81.5} &  \textbf{87.7} &  \textbf{92.7} &  \textbf{94.3} &  \textbf{90.8} &  \textbf{76.4} & \textbf{88.1} \\

\cline{2-15}          & \multicolumn{1}{c|}{\multirow{3}[2]{*}{BG\#1-2}} & GaitSet & 64.1  & 76.4  & 81.4  & 82.4  & 77.2  & 71.8  & 75.4  & 80.8  & 81.2  & 75.7  & 59.4  & 75.1  \\
&       & GaitGL  & 68.1 & 81.2 & 87.7 & 84.9 & 76.3 & 70.5 & 76.1 & 84.5 & 87.0 & 83.6 & 65.0 & 78.6 \\
&       & Ours  & \textbf{71.4} & \textbf{84.4} & \textbf{89.5} & \textbf{86.7} & \textbf{78.0} & \textbf{74.1} & \textbf{79.8} & \textbf{87.0} & \textbf{89.2} & \textbf{87.2} & \textbf{68.0} & \textbf{81.4} \\

\cline{2-15}          & \multicolumn{1}{c|}{\multirow{3}[2]{*}{CL\#1-2}} & GaitSet & 36.4  & 49.7  & 54.6  & 49.7  & 48.7  & 45.2  & 45.5  & 48.2  & 47.2  & 41.4  & 30.6  & 45.2  \\
    %
&       & GaitGL  & 46.9 & 58.7 & 66.6 & 65.4 & 58.3 & 54.1 & 59.5 & 62.7 & 61.3 & 57.1 & 40.6 & 57.4 \\
&       & Ours  & \textbf{46.9} & \textbf{64.7} & \textbf{71.8} & \textbf{69.2} & \textbf{63.7} & \textbf{58.0} & \textbf{64.4} & \textbf{68.4} & \textbf{65.9} & \textbf{60.3} & \textbf{41.1} & \textbf{61.3} \\
\hline


    \multicolumn{1}{c|}{\multirow{15}[2]{*}{\textbf{MT(62)}}} & \multicolumn{1}{c|}{\multirow{5}[2]{*}{NM\#5-6}} & AE    & 49.3  & 61.5  & 64.4  & 63.6  & 63.7  & 58.1  & 59.9  & 66.5  & 64.8  & 56.9  & 44.0  & 59.3  \\
          &       & MGAN  & 54.9  & 65.9  & 72.1  & 74.8  & 71.1  & 65.7  & 70.0  & 75.6  & 76.2  & 68.6  & 53.8  & 68.1  \\
          &       & GaitSet & 89.7  & 97.9  & 98.3  & 97.4  & 92.5  & 90.4  & 93.4  & 97.0  & 98.9  & 95.9  & 86.6  & 94.3  \\
&       & GaitGL  & \textbf{93.9} & 97.6 & 98.8 & \textbf{97.3} & \textbf{95.2} & 92.7 & \textbf{95.6} & 98.1 & 98.5 & \textbf{96.5} & \textbf{91.2} & 95.9 \\
&       & Ours  & 93.6 & \textbf{97.6} & \textbf{99.0} & 97.2 & 94.8 & \textbf{93.2} & 95.5 & \textbf{98.5} & \textbf{98.9} & 96.2 & 91.0 & \textbf{96.0} \\

\cline{2-15}          & \multicolumn{1}{c|}{\multirow{5}[2]{*}{BG\#1-2}} & AE    & 29.8  & 37.7  & 39.2  & 40.5  & 43.8  & 37.5  & 43.0  & 42.7  & 36.3  & 30.6  & 28.5  & 37.2  \\
          &       & MGAN  & 48.5  & 58.5  & 59.7  & 58.0  & 53.7  & 49.8  & 54.0  & 51.3  & 59.5  & 55.9  & 43.1  & 54.7  \\
         &       & GaitSet & 79.9  & 89.8  & 91.2  & 86.7  & 81.6  & 76.7  & 81.0  & 88.2  & 90.3  & 88.5  & 73.0  & 84.3  \\
&       & GaitGL  & 88.5 & \textbf{95.1} & 95.9 & 94.2 & 91.5 & 85.4 & 89.0 & 95.4 & \textbf{97.4} & 94.3 & \textbf{86.3} & 92.1 \\
&       & Ours  & \textbf{89.4} & 94.8 & \textbf{96.2} & \textbf{95.3} & \textbf{91.5} & \textbf{85.4} & \textbf{90.6} & \textbf{95.6} & 97.2 & \textbf{94.5} & 85.9 & \textbf{92.4} \\

\cline{2-15}          & \multicolumn{1}{c|}{\multirow{5}[2]{*}{CL\#1-2}} & AE    & 18.7  & 21.0  & 25.0  & 25.1  & 25.0  & 26.3  & 28.7  & 30.0  & 23.6  & 23.4  & 19.0  & 24.2  \\
          &       & MGAN  & 23.1  & 34.5  & 36.3  & 33.3  & 32.9  & 32.7  & 34.2  & 37.6  & 33.7  & 26.7  & 21.0  & 31.5  \\
         &       & GaitSet & 52.0  & 66.0  & 72.8  & 69.3  & 63.1  & 61.2  & 63.5  & 66.5  & 67.5  & 60.0  & 45.9  & 62.5  \\
&       & GaitGL  & \textbf{70.7} & 83.2 & 87.1 & 84.7 & 78.2 & 71.3 & {78.0} & {83.7} & 83.6 & 77.1 & \textbf{63.1} & 78.3 \\
&       & Ours  & 69.3 & \textbf{85.8} & \textbf{91.3} & \textbf{88.3} & \textbf{80.2} & \textbf{74.9} & \textbf{81.1} & \textbf{86.8} & \textbf{86.1} & \textbf{78.7} & 61.5 & \textbf{80.4} \\

    \hline

    \multicolumn{1}{c|}{\multirow{17}[2]{*}{\textbf{LT(74)}}} & \multicolumn{1}{c|}{\multirow{7}[2]{*}{NM\#5-6}} & CNN-3D & 87.1  & 93.2  & 97.0  & 94.6  & 90.2  & 88.3  & 91.1  & 93.8  & 96.5  & 96.0  & 85.7  & 92.1  \\
          &       & CNN-Ensemble & 88.7  & 95.1  & 98.2  & 96.4  & 94.1  & 91.5  & 93.9  & 97.5  & 98.4  & 95.8  & 85.6  & 94.1  \\
          &       & GaitSet & 91.1  & 99.0 & 99.9 & 97.8  & 95.1  & 94.5  & 96.1  & 98.3  & 99.2 & 98.1  & 88.0  & 96.1  \\
&       & ACL   & 92.0  & 98.5  & \textbf{100.0} & \textbf{98.9} & 95.7  & 91.5  & 94.5  & 97.7  & 98.4  & 96.7  & 91.9  & 96.0  \\
&       & GaitPart & 94.1  & 98.6 & 99.3  & 98.5  & 94.0  & 92.3  & 95.9  & 98.4  & 99.2 & 97.8  & 90.4  & 96.2  \\
&       & GaitGL  & 96.0 & 98.3  & 99.0  & 97.9  & \textbf{96.9} & 95.4 & 97.0 & 98.9 & 99.3 & 98.8 & 94.0 & 97.4 \\
&       & Ours  & \textbf{97.2} & \textbf{99.0}  & 99.4  & 97.8  & 96.8 & \textbf{96.1} & \textbf{97.9} & \textbf{99.9} & \textbf{99.8} & \textbf{99.5} & \textbf{94.8} & \textbf{98.0} \\

\cline{2-15}
          & \multicolumn{1}{c|}{\multirow{5}[2]{*}{BG\#1-2}} & CNN-LB & 64.2  & 80.6  & 82.7  & 76.9  & 64.8  & 63.1  & 68.0  & 76.9  & 82.2  & 75.4  & 61.3  & 72.4  \\
          &       & GaitSet & 86.7  & 94.2  & 95.7  & 93.4  & 88.9  & 85.5  & 89.0  & 91.7  & 94.5  & 95.9  & 83.3  & 90.8  \\
&       & GaitPart & 89.1  & 94.8 & 96.7 & 95.1 & 88.3  & 84.9 & 89.0  & 93.5  & 96.1  & 93.8  & 85.8  & 91.5  \\
&       & GaitGL  & 92.6 & 96.6 & 96.8 & 95.5 & 93.5 & 89.3 & 92.2 & 96.5 & 98.2 & 96.9 & 91.5 & 94.5 \\
&       & Ours  & \textbf{94.0} & \textbf{97.1} & \textbf{97.7} & \textbf{96.3} & \textbf{95.5} & \textbf{93.7} & \textbf{94.7} & \textbf{98.0} & \textbf{98.6} & \textbf{98.0} & \textbf{93.1} & \textbf{96.1} \\

\cline{2-15} & \multicolumn{1}{c|}{\multirow{5}[2]{*}{CL\#1-2}} & CNN-LB & 37.7  & 57.2  & 66.6  & 61.1  & 55.2  & 54.6  & 55.2  & 59.1  & 58.9  & 48.8  & 39.4  & 54.0  \\
          &       & GaitSet & 59.5  & 75.0  & 78.3  & 74.6  & 71.4  & 71.3  & 70.8  & 74.1  & 74.6  & 69.4  & 54.1  & 70.3  \\
&       & GaitPart & 70.7  & 85.5  & 86.9  & 83.3  & 77.1  & 72.5  & 76.9  & 82.2  & 83.8  & 80.2  & 66.5  & 78.7  \\
&       & GaitGL  & 76.6 & 90.0 & 90.3 & 87.1 & 84.5 & 79.0 & 84.1 & 87.0 & 87.3 & 84.4 & 69.5 & 83.6 \\
&       & Ours  & \textbf{78.5} & \textbf{92.6} & \textbf{94.1} & \textbf{91.4} & \textbf{86.9} & \textbf{82.9} & \textbf{88.4} & \textbf{90.9} & \textbf{91.5} & \textbf{87.7} & \textbf{72.5} & \textbf{87.0} \\
    \bottomrule
    \end{tabular}%
\label{tab:sota_casiab}
\vspace*{-2em}
\end{table*}%
\textbf{Ablation Study on Noisy CASIA-B} Ablation results are shown in Table.\ref{tab:ablation}. We categorize the experiments into supervised and self-supervised by the loss. $L_s$ denotes the supervised loss including CE and triplet. Note that for fair comparison, all methods are using the same hyper parameters. Following conclusions can be safely made. 1) Experiments \#1 and \#2, \#4 and \#5, \#6 and \#7 demonstrate that the cyclic training in Alg.\ref{alg} is effective for relieving memorization effects and improving performances. 2) Experiments \#2 and \#3, \#3 and \#8 demonstrates the effectiveness of AND module in noisy settings. 3)  Experiments \#4 and \#5 show that pure self-supervised is not practical in this framework. 4)  Experiments \#1 and \#6,  \#2 and \#7,  \#3 and \#8 verify that the CRC term brings an improvement consistantly. Although $L_c$ itself does not work in this framework, adding a term $L_c$ to the framework encourages the two networks to learn from each other and correct each other's memorization mistakes. 5) Experiments \#1 and \#8 show that CNTN is able to address clothing and noisy labels memorization in gait recognition.
\\\textbf{Outdoor-Gait} Experiments results show that nearly all metrics increase compared to the baseline. The table and more details are in Appendix A.

\subsection{Experiments on Benchmarks}
Since traditional benchmarks inevitably contain noisy data, CNTN addresses memorization effect and thus the experiments show CNTN's effectiveness.
\\\textbf{Backbone} GaitGL \cite{lin2021gait} is set as our backbone, which utilizes 3D convolutional network and multi-scale feature extractors to get refined representations.
\\\textbf{CASIA-B} In Tab.\ref{tab:sota_casiab}, CNTN is compared with state-of-the-art gait recognition approaches on CASIA-B, including
GaitSet \cite{chao2019gaitset}, AE \cite{yu2017invariant}, MGAN \cite{he2018multi}, CNN-LB, CNN-3D, CNN-Ensemble \cite{wu2016comprehensive}, ACL \cite{zhang2019cross}, GaitPart \cite{fan2020gaitpart} and GaitGL \cite{lin2021gait}. There are following observations. 1) The proposed method achieves the best recognition accuracy at all mean accuracy, and in most view cases. 2) More importantly, the proposed method surpasses other methods with a great margin in CL scenario, which demonstrates CNTN alleviates the appearance memorization effect.  
\\\textbf{Outdoor-Gait} In Appendix A Tab.\ref{table:outdoors}, CNTN is validated on Outdoor-Gait, and previous methods include GEI+PCA \cite{HOFMANN2014195}, GEI+Net \cite{shiraga2016geinet}, GaitNet \cite{song2019gaitnet}. Note that we adopt GaitSet as the baseline, and the results outperform state-of-the-art methods by a large margin. 
\\\textbf{OUMVLP} In Tab.\ref{comparision_oumvlp}, CNTN is conducted on OUMVLP, and other methods include GEINet \cite{shiraga2016geinet}, GaitSet \cite{chao2019gaitset}, GaitPart \cite{fan2020gaitpart}, GLN \cite{hou2020gait}, GaitGL \cite{lin2021gait}. The experiment results show that the proposed method can achieve state-of-the-art results in all views.
\\\textbf{Model Agnostic Results} In Tab.\ref{tab:agnostic}, note that the inference model is only one stream of the trained model (F), thus the model parameters stay the same as the original backbone. The results show that CNTN improves the performances of the original baseline at all metrics consistently, regardless of backbones.


\section{Conclusion}
In this work, to better equip the model with ability to resist memorization effect, we propose a cyclic
noise-tolerant network that integrates forgetting network and memorizing network, the algorithm and the co-teaching robust constraint into a unified method. The alternating cyclic learning between the two networks can effectively reduce the influence of the noisy samples. Compared to other co-teaching methods, CNTN is cyclic and explicitly makes the two networks with different abilities.
Furthermore, our framework is compatible with any neural network and can be applicable for other noisy learning tasks. Experiments on three gait recognition benchmarks and two reconstructed noisy gait recognition datasets demonstrate the effectiveness of the proposed algorithm and achieves state-of-the-art performance.
\clearpage
\bibliography{egbib}

\clearpage
\appendix

 \begin{figure*}[ht]
	\centering
	\includegraphics[width=\textwidth]{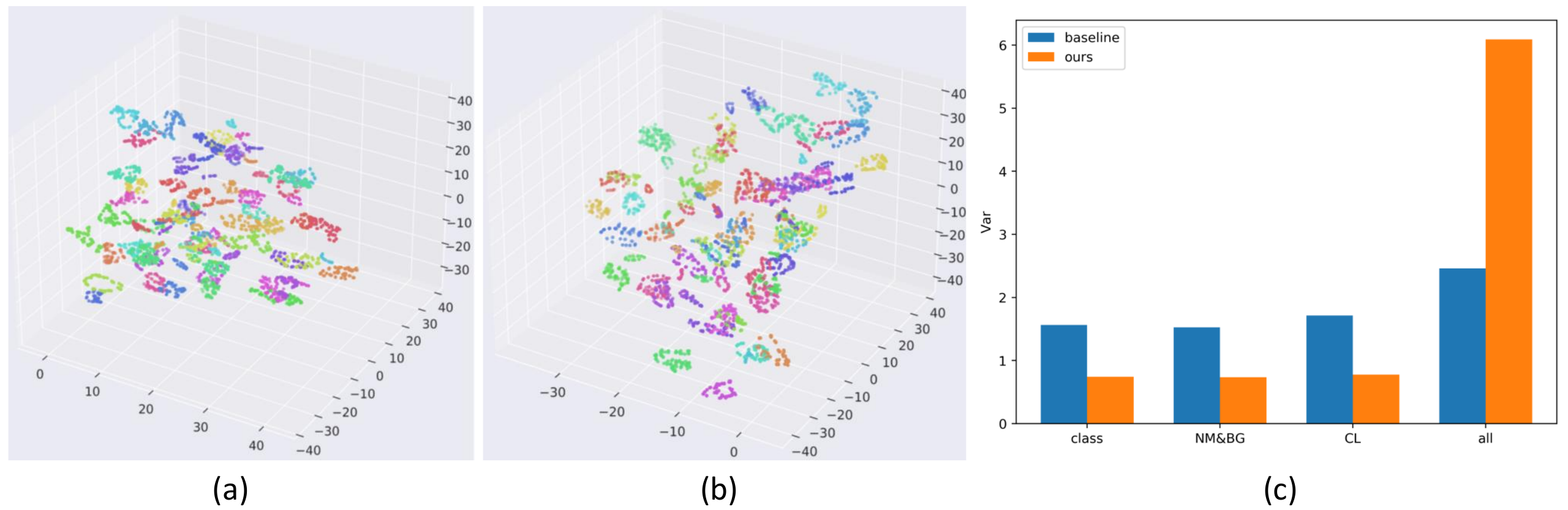}
	\caption{Analysis on test set. (a) The TSNE visualization of baseline features; (b) The TSNE visualization of CNTN features; (c) Average variance of features; `class' denotes the average variance of features from the same class; `NM\&BG' denotes the average variance of features from the same class and the walking conditions in NM and BG; `CL' denotes the average variance of features from the same class and the walking conditions in CL; `all' denotes the variance of all test features.}
	\label{fig:vis}
\end{figure*}

 \begin{figure}[ht]
	\centering
\includegraphics[width=0.45\textwidth]{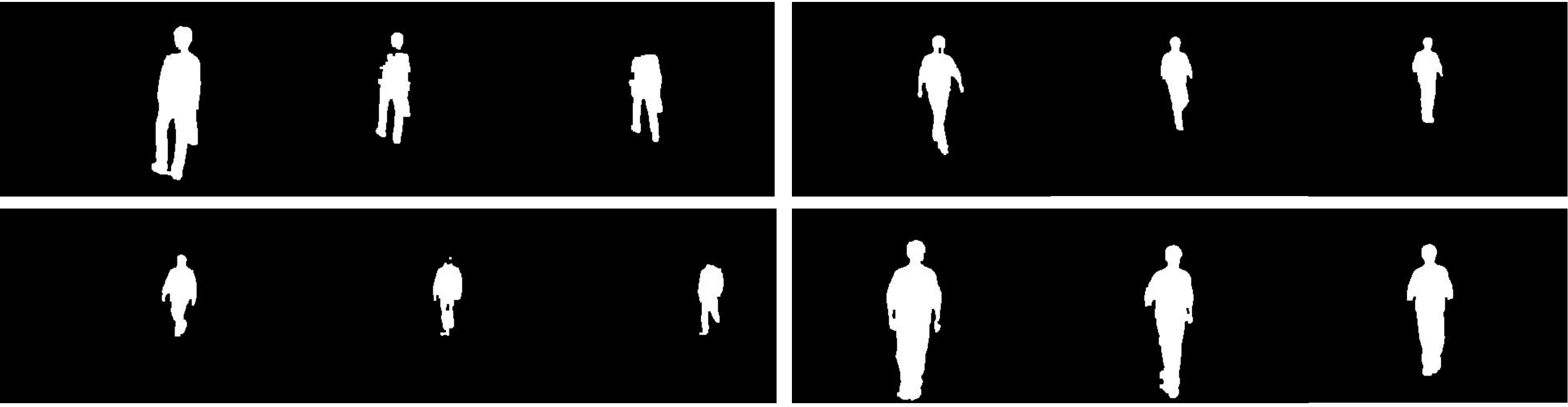}
	\caption{Noise in existing gait silhouette benchmarks. Each
pair of images contain the same identity from the same view and in the same walking condition, with the left being an inlier, the middle with some noise, and the right with large noise.}
	\label{fig:silh}
\end{figure}

\begin{table*}[ht]
	\begin{center}
		\centering
		\caption{
			The hyperparameters used in our experiments. BS, LR, OPT, GAMA are the batch size, learning rate, optimizer and the learning rate decay ration.  $\sigma_{0}$-$\sigma_{2}$ are the coefficients of supervised and co-teaching losses.
			Milestone means the iteration of learning rate decay and Iters means the total training iteration.
		}
		\label{tab:paras}
		\setlength{\tabcolsep}{1.3mm}
		{
			\begin{tabular}{ c | c | c | c | c | c | c | c | c | c }
				\toprule[1.2pt]
				Backbone  & BS  & LR & OPT & GAMA & $\sigma_{0}$ & $\sigma_{1}$ & $\sigma_{2}$  & Milestone & Iters    \\
				\hline
				GaitSet  & [8, 16] & 1e-1 & SGD & 0.1 & 0.1 & 1.0 & 0.1 & [10k, 20k]  & 20k  \\
				GaitGL  & [8, 8] & 1e-4  & ADAM & 0.1 & 0.1 & 1.0 & 0.1 & [70k]  & 80k   \\
				\bottomrule[1.2pt]
			\end{tabular}
		}
	\end{center}
\end{table*}

\begin{table*}[ht]
		\renewcommand{\arraystretch}{1.3}
		\centering
		\caption{Rank-1 accuracy (\%) on Noisy Clothing Outdoor-Gait, different settings and conditions.}
		\setlength{\tabcolsep}{2.0mm}
		{
			\begin{threeparttable}[b]
				\begin{tabular}{c|c|c|c|c|c|c|c|c|c|c}
        			\hline
					Gallery & \multicolumn{3}{c|}{\textbf{NM}} & \multicolumn{3}{c|}{\textbf{CL}} & \multicolumn{3}{c|}{\textbf{BG}} & \multirow{2}*{\textbf{ Mean}} \\
					\cline{2-10} Probe & NM & CL & BG & NM & CL & BG & NM & CL & BG &  \\
        			\hline
					 Baseline &96.7 &87.2 &93.1 &\textbf{85.0} &96.9 &79.9  & 90.6 & 82.6 & 96.1 &89.8  \\
					 \hline
					\textbf{Ours} &\textbf{98.2} &\textbf{90.0} &\textbf{94.7} &{84.9} &\textbf{98.2} & \textbf{82.3} &  \textbf{93.6} & \textbf{85.6} & \textbf{97.5} & \textbf{91.7} \\
        			\hline
				\end{tabular}
			\end{threeparttable}
		}

		\label{table:noisy_outdoors}
	\end{table*}

 \begin{table*}[ht]
		\renewcommand{\arraystretch}{1.3}
		\centering
		\caption{Rank-1 accuracy (\%) on Outdoor-Gait, different settings and conditions.}
		\setlength{\tabcolsep}{2.0mm}
		{
			\begin{threeparttable}[b]
				\begin{tabular}{c|c|c|c|c|c|c|c|c|c|c}
        			\hline

					Gallery & \multicolumn{3}{c|}{\textbf{NM}} & \multicolumn{3}{c|}{\textbf{CL}} & \multicolumn{3}{c|}{\textbf{BG}} & \multirow{2}*{\textbf{ Mean}} \\
					\cline{2-10} Probe & NM & CL & BG & NM & CL & BG & NM & CL & BG &  \\
        			\hline

					GEI+PCA &85.0 &29.5 &38.9 &29.2 &86.7 & 22.9 & 30.7 & 20.8 & 92.5 & 48.5 \\
					        			\hline
					GEI+Net &93.2 &55.8 &59.2 &45.9 &93.7 & 36.7 & 44.2 & 27.5 & 96.6 & 48.5 \\
					        			\hline
					GaitNet &96.9 &60.2 &89.1 &58.7 &\textbf{97.3} & 55.4 & {92.0} & 59.7 & \textbf{97.1} & 78.5 \\
					\hline
					\textbf{baseline} &{96.4} &{90.3} &{92.3} &{91.3} &{96.8} & {87.8}  &{91.1} &{85.7} & {95.5}  & {94.4}\\
					 \hline
					\textbf{Ours} &\textbf{97.4} &\textbf{91.9} &\textbf{94.0} &\textbf{91.6} &{97.1} & \textbf{88.5}  & \textbf{92.8} & \textbf{86.4} & {96.7}  & \textbf{94.9}\\
        			\hline
				\end{tabular}
			\end{threeparttable}
		}

		\label{table:outdoors}
	\end{table*}

\section{Experiments}
 \subsection{Datasets}
\textbf{CASIA-B}  The CASIA-B dataset \cite{6115889} is the popular cross-cloth gait database. It includes 124 subjects, each of which has 10 groups of videos. Among these groups, six of them are sampled in normal walking (NM), two groups are in
walking with a bag (BG), and the rest are in walking in different cloth (CL). Each group contains 11 gait sequences from different angles ($0^{\circ}$-$180^{\circ}$
and the sampling interval is $18^{\circ}$). Therefore, there are 124 (subject) × 10 (groups) × 11 (view
angle) = 13,640 gait sequences in CASIA-B. The gait sequence of each subject are divided into training set and
test set. Following the setting of previous works \cite{chao2021gaitset}, in small-sample training (ST) the first
24 subjects (labeled in 001-024) are used for training and the
rest 100 subjects are leaved for test. In medium-sample
training (MT), the first 62 subjects are used for training and the rest 62 subjects are leaved for test. In large-sample training (LT), the first 74 subjects are used for training and
the rest 50 subjects are leaved for test. In the test stage,
the sequences NM\#01-NM\#04 are taken as the gallery set,
while the sequences NM\#05-NM\#06, BG\#01-BG\#02, and
CL\#01-CL\#02 are considered as the probe set to evaluate
the performance.

\textbf{Noisy-CASIA-B} We construct Noisy-CASIA-B into two different settings. The first is normal noise setting with noise rate 0.1 and 0.2 which contains random noise and appearance noise. Random noise corresponds to the random label noise, which indicates that the noise percentage of original gait sequences are labeled with random noisy labels in this dataset, regardless of its walking conditions (eg, '001 NM\#01' $\rightarrow$ '002 NM\#01', '001 CL\#01' $\rightarrow$ '002 CL\#01'). Appearance noise corresponds to the appearance random disturbance which simulates the clothing change, containing morphological operations. The second is noisy clothing setting where one pedestrian is separated to two identities and creates a new identity (eg, '001 NM\#01' $\rightarrow$ '001NM\#01', '001 CL\#01' $\rightarrow$ '130NM\#01'), with a percentage of 0.6 (In train set, the whole dataset is '001'-'074', noisy ones are '001'-'044' ). Noisy Clothing CASIA-B is the most practical setting for gait recognition since in industry, it is usually the pedestrians with different clothing that makes both annotators and unsupervised clustering methods confused and thus with incorrect labels.
\\\textbf{Outdoor-Gait}. The Outdoor-Gait \cite{song2019gaitnet} dataset also contains rich clothing and bag variations with complex outdoor backgrounds. Outdoor-Gait contains 138 people with 3 different clothing conditions (NM: normal, CL: different cloth, BG: with bag) in 3 scenes
(SCENE-1: simple background, SCENE-2: static and complex background, SCENE-3: dynamic and complex background with moving objects).  Following the setting of previous works \cite{chao2019gaitset}, we take the first 69 subjects for training and the left subjects for testing and for each condition, there are
at least 2 video sequences in gallery and probe.
\\\textbf{Noisy-Outdoor-Gait} We construct Noisy-Outdoor-Gait which accords with Noisy-CASIA-B.
\\\textbf{OUMVLP} OUMVLP \cite{takemura2018multi} is one of the largest view-variation gait dataset. This dataset includes more than 10,000 subjects. Each subject’s sequences are captured under 14 views ($0^{\circ}$-$270^{\circ}$ and the sampling interval is $15^{\circ}$). There are two sequences under each view. There is no CL or BG walking condition.
\\\textbf{Evaluation metrics}
In the testing phase, we compare the feature similarities between probe and gallery samples to identify a person and report performance of the average Rank-1 recognition accuracy.
\label{sec:settings}
\subsection{Noisy Settings} The experiment settings in this paper are twofold, including experiments on three benchmarks and experiments on two reconstructed noisy datasets. Experiments on three benchmarks just follows everything the same in previous works \cite{lin2020gait,chao2021gaitset}, and the results of CASIA-B, Outdoor-Gait and OUMVLP are in Tab.\ref{tab:sota_casiab}, Tab.\ref{table:outdoors} and Tab.\ref{comparision_oumvlp} respectively. Experiments on noisy settings includes normal noise setting and noisy clothing setting. Normal noise setting follows the noise setting in other noisy tasks (noisy image classification on CIFAR10 and CIFAR100) \cite{han2018co,yu2019does}, and results are in Tab.\ref{tab:random casiab}. However, this random noise setting is not close to practical situations in gait recognition task since the annotators and the unsupervised clustering methods have a clear tendency that the pedestrians with different clothing are misclassified. Thus, to better simulate the practical clothing noise, we also includes a noisy clothing setting, and the results of Noisy Clothing CASIA-B and Noisy Clothing Outdoor-Gait are in Tab.\ref{tab:noisy} and Tab.\ref{table:noisy_outdoors} respectively.


\subsection{Visualization Feature Representations on Noisy-CASIA-B}
Although TSNE is just one way to compact the features to visualized dimensions, it can still provide us with some intuitional feelings.
In Fig.\ref{fig:vis}, CNTN learns a larger space than baseline, while at the same time, every class cluster is more compact and aggregated. Fig.\ref{fig:vis} (c), the variance on the test set can also support this. The average class variance nearly halves the baseline, while the total variance triples the baseline. The larger space and better feature distribution is one of the reflections that the CNTN learns a pattern with less memorization and better generalization.

\subsection{Results on Outdoor-Gait}
Experiments results of outdoor-Gait dataset are shown in Table.\ref{table:outdoors}. The baseline is GaitSet \cite{chao2019gaitset,chao2021gaitset}. The results show that nearly all metrics increase compared to the baseline.

\subsection{Results on Noisy-Outdoor-Gait}
Experiments results of Noisy outdoor-Gait dataset are shown in Table.\ref{table:noisy_outdoors}. The baseline is GaitSet \cite{chao2019gaitset,chao2021gaitset}. The results show that nearly all metrics increase compared to the baseline. And the average accuracy gains about 2\%, which shows the effectiveness of CNTN.

\subsection{Implementation details.}
In Table. \ref{tab:paras}, we elaborate the setting of the hyperparameters used in CASIA-B on different models. Our baseline encoder is GaitSet~\cite{chao2019gaitset} and the other settings are the same as \cite{chao2019gaitset,lin2021gait}. The EMA ratio is set to 0.99 in all our experiments.
We adopt the same preprocessing approach as \cite{chao2019gaitset} to obtain gait silhouettes for CASIA-B and Outdoor-gait. \\
\textbf{CASIA-B} In the setting of noisy experiments,  the backbone network is GaitSet and the iteration number is set to 20K. The number of subjects and the number of sequences for each subject are set to (8, 16) and the input sizes is (64, 44). For evaluation, all silhouettes of gait sequences are taken to obtain the final representation. We use the loss function coefficients weight decay to guarantee better convergence. Specifically, $\sigma_{0}$  and $\sigma_{2}$  gradually increase from 0.01 to 0.1, and $\sigma_{2}$  gradually decrease from 1 to 0.1.
Regarding the state-of-the-art experiments, we do not use the loss function coefficients weight decay.
We implement our method on the basis of OpenGait\footnote{\href{https://github.com/ShiqiYu/OpenGait.git}{https://github.com/ShiqiYu/OpenGait.git}}. For a fair comparison, we use the same settings in OpenGait.
\\\textbf{Outdoor-Gait} In the setting of all experiments, the backbone network is GaitSet and the iteration number is set to 20K. Other settings are the same as Table. \ref{tab:paras}.
\\\textbf{OUMVLP} In the setting of all experiments, the backbone network is GaitGL and the iteration number is set to 80K. Other settings are the same as Table. \ref{tab:paras}.

\end{document}